\documentclass{article}

\pdfoutput=1
\PassOptionsToPackage{numbers, compress}{natbib}

\usepackage[preprint]{neurips_2024}

\usepackage[utf8]{inputenc} 
\usepackage[T1]{fontenc}    
\usepackage{hyperref}       
\usepackage{url}            
\usepackage{booktabs}       
\usepackage{amsfonts}       
\usepackage{nicefrac}       
\usepackage{microtype}      
\usepackage{graphicx}
\usepackage{wrapfig}
\usepackage{mathrsfs}
\usepackage{amssymb}
\usepackage{algorithm}
\usepackage{algorithmic}
\usepackage[algo2e]{algorithm2e}
\usepackage{multirow}
\usepackage{float}
\usepackage{tabularx}
\usepackage{booktabs}
\usepackage{subcaption}

\usepackage{alltt}
\usepackage[dvipsnames]{xcolor,colortbl}          
\definecolor{lightgray}{gray}{0.9}
\definecolor{linecolor}{rgb}{0.82, 0.94, 0.75}
\definecolor{citecolor}{HTML}{0071BC}
\definecolor{linkcolor}{HTML}{ED1C24}

\definecolor{Image}{rgb}{0.21,0.49,0.74}
\definecolor{Point Cloud}{RGB}{105,145,60}

\usepackage{amsmath}

\usepackage{cleveref}

\SetKwComment{Comment}{$\triangleright$\ }{}
\newcommand{\methodshort}[1]{\textsc{Adapter-X}}

\newcommand{\ie}{\emph{i.e.}}

\definecolor{evaunit01green}{RGB}{82,208,83}
\definecolor{lowred}{RGB}{238,18,137}

\definecolor{lowerred}{RGB}{255,110,180}

\newcommand{\dplus}[1]{\fontsize{6pt}{0.1em}\selectfont (\textbf{\textcolor{lowred}{#1}})}
\newcommand{\ddplus}[1]{\fontsize{6pt}{0.1em}\selectfont (\textbf{\textcolor{lowerred}{#1}})}

\definecolor{rowunit}{RGB}{128,128,255}
\definecolor{evaunit01green}{RGB}{54,125,189}
\newcommand{\evagreen}[1]{\textcolor{evaunit01green}{#1}}
\newcommand{\dtplus}[1]{\fontsize{6pt}{0.1em}\selectfont (\textbf{\evagreen{#1}})}

\definecolor{lightgray}{rgb}{0.9, 0.9, 0.9}
\definecolor{ballblue}{rgb}{0.13, 0.67, 0.8}
\newcommand{\STAB}[1]{\begin{tabular}{@{}c@{}}#1\end{tabular}}

\title{\methodshort{}: A Novel General Parameter-Efficient Fine-Tuning Framework for Vision}

\author{%
Minglei Li$^{1\dagger}$ \quad
Peng Ye$^{1\dagger}$ \quad
Yongqi Huang$^1$ \quad
Lin Zhang$^1$ \quad
\\
\textbf{Tao Chen}$^1$\thanks{\vspace{-15pt}Corresponding Author. ~~~$^\dagger$Equal Contribution.} \quad
\textbf{Tong He}$^2$ \quad
\textbf{Jiayuan Fan}$^1$ \quad
\textbf{Wanli Ouyang}$^2$ \quad
\\
{\tt\small mlli23@m.fudan.edu.cn, eetchen@fudan.edu.cn}
\\
$^1$Fudan University \quad $^2$ Shanghai AI Laboratory
\\{\tt\small \textbf{\href{https://github.com/leoli646/Adapter-X}{https://github.com/leoli646/Adapter-X}}}
}

\begin{document}

\maketitle

\begin{abstract}
  Parameter-efficient fine-tuning (PEFT) has become increasingly important as foundation models continue to grow in both popularity and size. Adapter has been particularly well-received 
  due to their potential for parameter reduction and adaptability across diverse tasks.
  However, striking a balance between high efficiency and robust generalization across tasks remains a challenge for adapter-based methods.  
  We analyze existing methods and find that: 1) parameter sharing is the key to reducing redundancy; 2) more tunable parameters, dynamic allocation, and block-specific design are keys to improving performance. Unfortunately, no previous work considers all these factors. Inspired by this insight, we introduce a novel framework named \methodshort{}. First, a Sharing Mixture of Adapters (SMoA) module is proposed to fulfill token-level dynamic allocation, increased tunable parameters, and inter-block sharing at the same time.
  Second, some block-specific designs like Prompt Generator (PG) are introduced to further enhance the ability of adaptation.
  Extensive experiments across 2D image and 3D point cloud modalities demonstrate that Adapter-X represents a significant milestone as it is the first to outperform full fine-tuning 
  in both 2D image and 3D point cloud modalities with significantly fewer parameters, 
  i.e., only $0.20\%$ and $1.88\%$ of original trainable parameters for 2D and 3D classification tasks. Our code will be publicly available.

\end{abstract}
\section{Introduction}
\label{sec:introduction}

In an era where models are becoming increasingly larger and more complex, the ability to efficiently adapt them to specific tasks and datasets has become increasingly crucial. Parameter-efficient fine-tuning (PEFT) has emerged as a promising solution to this challenge. Among the techniques proposed for PEFT, adapter-based methods have attracted significant interest due to the few trainable parameters and the potential to enhance the model's performance on different downstream tasks.

\begin{figure}
\centering
\includegraphics[width=1.0\linewidth]{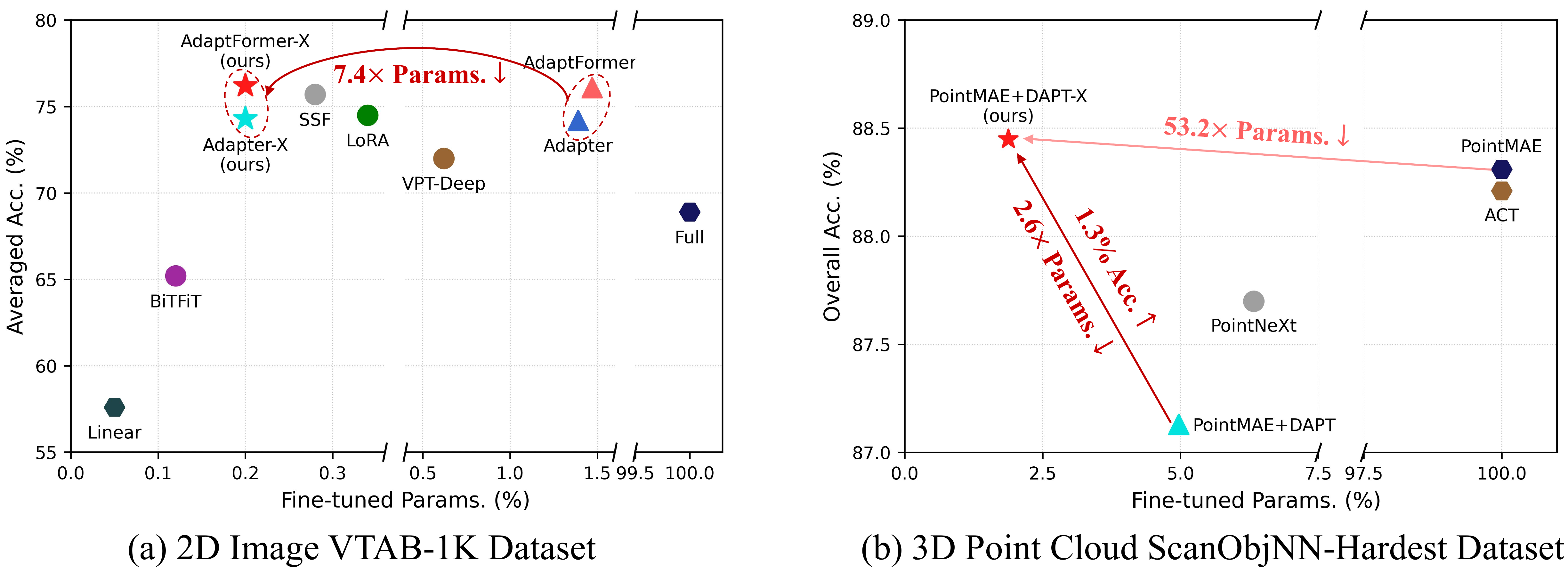}
\caption{Comparison of efficiency and performance between our \methodshort{} and other methods. }
\label{fig:perf_compare}
\vspace{-0.5cm}
\end{figure}

Adapter-based methods can primarily be organized into two categories, focusing on two distinct advantages: the performance capabilities across various tasks and the efficiency gained from the minimal scale of parameters required for fine-tuning.
Methods focusing on performance improvements are represented by three distinct approaches.~\cite{pfeiffer-2020-adapterfusion}\cite{he2021towards} aim to create a versatile adapter structure capable of handling a broad spectrum of tasks.~\cite{mao2022unipelt}~\cite{chen2023parameter}~\cite{he2021towards}~\cite{hu2023llm} aim to either combine the advantages of diverse PEFT approaches or seek to establish a unified perspective by analyzing the similarities among different adapters and other PEFT methods.~\cite{noah}~\cite{zhou2024autopeft} utilize neural architecture search (NAS) to discover superior PEFT combination approaches.
For the aspect of adapter efficiency, specific methods~\cite{pevit}~\cite{compactor}~\cite{sparseadapter}~\cite{jie2023revisiting} have been devised. They further improve adapter efficiency from different perspectives of redundancy, including rank, density, and precision.
While these methods strive to balance the dual advantages of adapters, none have yet achieved both high efficiency and good generalization across various modalities and tasks,
which invites us to ponder: \emph{How to further 
enhance generalization capabilities of Adapters while inspiring their efficiency potential?}

To answer this question, we analyze existing methods and summarize the following factors influencing efficiency and performance. 
From an efficiency standpoint, 
lots of approaches~\cite{compactor}~\cite{pevit}~\cite{kopiczko2023vera}~\cite{renduchintala2023tied}~\cite{wang2024prolora} have employed a sharing strategy between different blocks, which show that the parameter-sharing mechanism can indeed reduce redundancy, thus enhancing the parameter efficiency of PEFT methods.
From a performance perspective, we find that improving the performance of the existing PEFT methods on different tasks can be roughly categorized into three primary perspectives. 
Firstly, increasing the trainable parameters, such as increasing the rank of adapters, can enhance the adapter's performance. 
Secondly, dynamic allocation of tunable parameters has proven to be highly beneficial in the context of PEFT. 
For instance, the NAS-based strategy~\cite{noah}~\cite{zhou2024autopeft} can be considered as allocating parameters following the change of data distribution.
Thirdly, diverse methods implement distinct designs for blocks at different positions within the architecture, such as utilizing different PEFT methods. These block-specific designs enable the model to better accommodate a range of datasets.
Previous works only focus on one or a few key points, making them unable to handle the balance problem well.
We suppose that if we can design a method that can consider all these effective factors at the same time, we can achieve concurrent improvements in both efficiency and performance.

To this end, we propose a new general parameter-efficient fine-tuning framework named \methodshort{}.
The core of the framework is that different blocks recursively select the most suitable adapter from a block sharing adapter expert library. We refer to this component as the Sharing Mixture of Adapters (SMoA). 
This element endows us with the three aforementioned advantages: sharing strategy, dynamic allocation, and an increased number of tunable parameters for each block. 
First, Sharing Mixture of Adapters, is utilized across all blocks, significantly reducing the parameter count. 
Secondly, for any given set of tokens, \methodshort{} will choose distinct adapter experts for different blocks to process them, thereby achieving a token-level dynamic allocation of fine-tuning parameters.
Third, since each block will recursively access all the adapter experts from the shared expert library, it implies that each block can access more adapters than the previous traditional approach.
Lastly, in order to enhance the block-specific design, \methodshort{} utilizes different norm layers and Prompt Generator to generate block-specific prompts, further improving the model's performance.

In order to validate the effectiveness and generalization capability of the proposed method \methodshort{}, extensive experiments are conducted on different datasets across 2D image and 3D point cloud modalities. The results demonstrate the ability of \methodshort{} to achieve superior performance with a significantly reduced number of parameters, as shown in Fig.~\ref{fig:perf_compare}.
To the best of our knowledge, this is the first work that outperforms full fine-tuning in both  2D image and 3D point cloud modalities. In particular, on the 2D VTAB dataset, our method achieves better classification results with 7.5 times less parameters compared to the original Adapter approach. On the 3D ScanObjectNN dataset, our method performs up to $1.32\%$ better than the SOTA PEFT approach with less than half of the tunable parameters of it. Our contributions can be summarized as: 
\begin{itemize}
    \item We first analyze existing PEFT works and summarize four factors related to improving efficiency and performance: sharing strategy, larger tunable parameters count, dynamic allocation, and block-specific design.
    \item Based on the above factors, we design a novel general parameter-efficient fine-tuning framework named \methodshort{}. A Sharing  Mixture of Adapters (SMoA) is proposed to dynamically share adapters, which in turn increases parameters each block can access. We further adopt the idea of block-specific design and introduce different norm layer and generated prompt for each block, further enhancing the performance.
    \item Our proposed \methodshort{} is the first to outperform full fine-tuning in both 2D image and 3D point cloud modalities, even performing better than the SOTA PEFT method tailored for point cloud modality with less tunable parameters.
\end{itemize}

\section{Related Works}
\label{sec:related_work}

\textbf{Adapter Series}, as introduced in \cite{houlsby2019parameter}, are lightweight modules that enhance Transformer layers with a structure comprising a linear down-projector, a nonlinear activation, an up-projector, and a residual connection. 
Through the continuous improvement of ~\cite{pfeiffer-2020-adapterfusion}\cite{he2021towards}, the current mainstream Adapter usage method\cite{zhu2021counter}\cite{lei2024conditional}\cite{edalati2022krona} is to add a parallel original Adapter at the FFN position of each block, and scale the output.
Existing methods aim to improve the adapter's performance on specific tasks by introducing specialized designs for individual blocks. For instance, approaches like ~\cite{mao2022unipelt}~\cite{chen2023parameter}~\cite{he2021towards}~\cite{hu2023llm} either activate various PEFT techniques, such as LoRA~\cite{lora}, prefix-tuning~\cite{li2021prefix}, within distinct blocks or blend their outputs in varying ratios. Furthermore, methods like ~\cite{noah}~\cite{zhou2024autopeft} utilize NAS to discover a range of PEFT configurations that are superior for different tasks.
Besides, some work focuses on more compact designs for adapters. ~\cite{compactor}~\cite{pevit} regard the weights of adapters as the Kronecker product of two smaller matrices, one of which is shared among adapters. ~\cite{sparseadapter} prunes the dense weights of adapters before fine-tuning.
In the domain of computer vision, for 2D image processing tasks, ~\cite{chen2022adaptformer} employs a scaling mechanism in conjunction with Adapter to adapt different visual tasks. ~\cite{noah} leverages search methodologies to identify different configurations across blocks for adaptation. ~\cite{jie2023revisiting} improves parameter efficiency of Adapters with quantization from the perspective of precision redundancy. 
When it comes to 3D point cloud, ~\cite{zha2023instance} extends the Prompt tuning with a DGCNN~\cite{phan2018dgcnn} to extract instance-aware prompts for model fine-tuning instead of using static prompts. DAPT~\cite{zhou2024dapt} proposes Dynamic Adapter and integrates it with Prompt Tuning to reduce the tunable parameters and achieve remarkable performance. 
While various strategies are used to decrease trainable parameters, our method of sharing adapters across blocks offers a distinct advantage on efficient adaptation on various tasks. Additionally, it is easy to achieve integration with other advanced adapter method.

\textbf{Parameter Sharing}, as a method to reduce model size and computational costs, has been utilized in various studies. 
For instance, ~\cite{Dehghani2018}\cite{Takase2023}~\cite{Lou2021}\cite{Reid2021} explores strategies for sharing parameters between different layers of the Transformer to reduce the number of trainable parameters and increase model calculation speed. 
~\cite{Ge2022}~\cite{Pires2023} share a complete module of Transformer to reduce the parameter count.
Recent advancements have taken this a step further by combining parameter sharing with PEFT methods. VERA~\cite{kopiczko2023vera} shares two frozen random matrices across all layers, updating disentangled combination vectors for each layer, while Tied-LoRA~\cite{renduchintala2023tied} enhances efficiency through weight tying and selective training.
PRoLoRA~\cite{wang2024prolora} introduces an intra-layer sharing mechanism which greatly reduces the number of trainable parameters.
Instead of simply sharing the same fine-tuning parameters for different blocks to reduce trainable parameters, our proposed method, \methodshort{}, realizes fine-tuning parameters allocation from block-general level with sharing mechanism which we find crucial to Adapters. 

\textbf{Mixture of Experts(MoE)} is an ensemble method, typically envisioned as a group of specialized sub-modules, known as experts, each adept at handling distinct types of input data. These experts are controlled by a router that dynamically selects the most relevant expert based on the input.  
Inspired by the ideas of MoE and its advanced variants MH-MoE~\cite{wu2024multi} and X-MoE~\cite{chi2022representation}, we first redefine the role of standard adapter as expert within our framework, and introduce a more efficient and effective procedure of processing input to select a more compatible adapter for each block. 
\section{Method}
\label{sec:method}

\begin{figure*}

\centering
\includegraphics[width=1.0\linewidth]{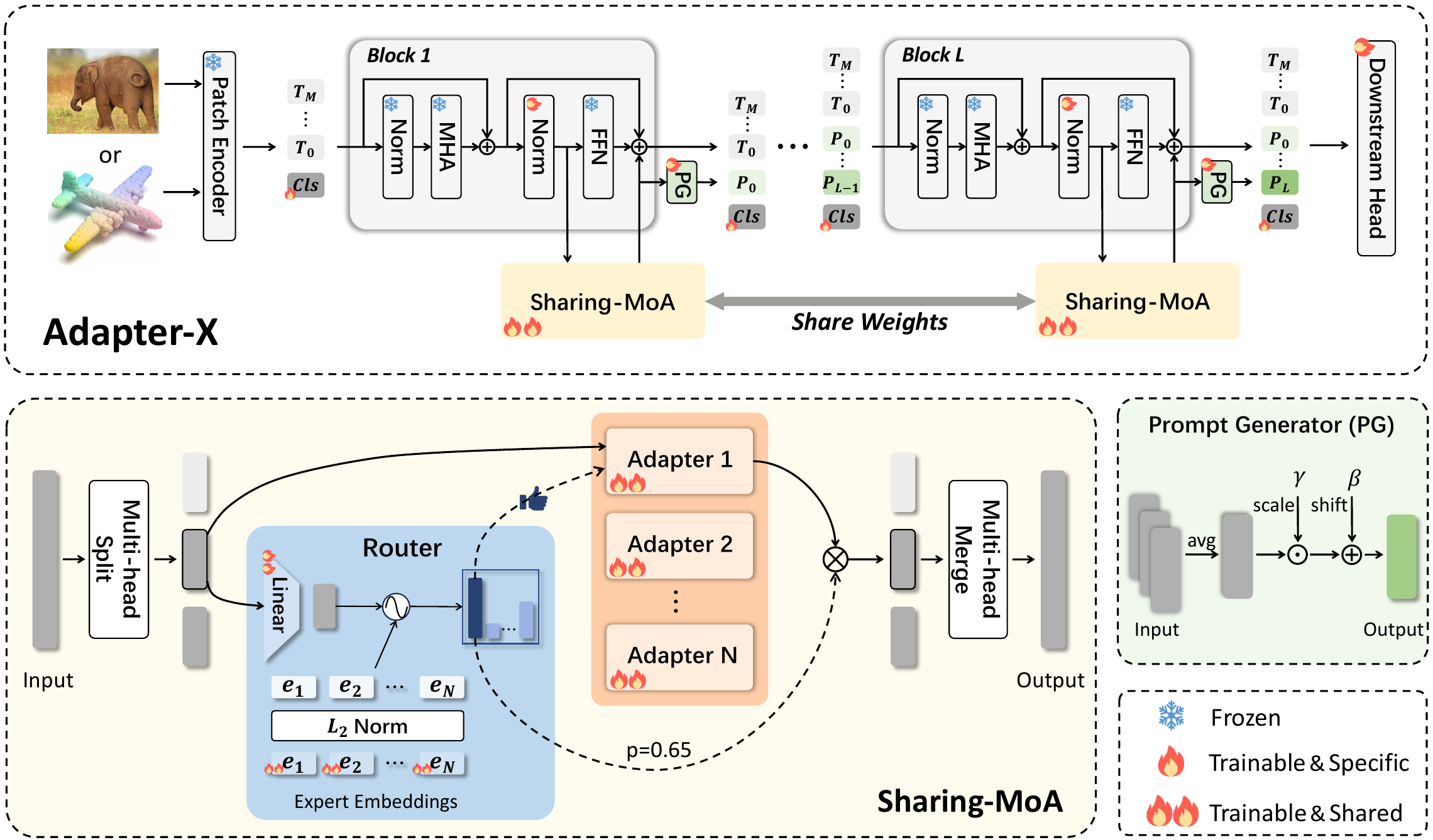}
\caption{The overview of the proposed \methodshort{}. By comprehensively considering parameter sharing, more tunable parameters, dynamic allocation, and block-specific design, the model achieves efficient adaptation to different tasks with minimal trainable parameters.}
\label{fig:AdapterX_framework}
\vspace{-0.3cm}
\end{figure*}

The overview of \methodshort{} is depicted in Fig. \ref{fig:AdapterX_framework}. In this following, we first give a brief review of Adapter and MoE in Sec. \ref{sec:Pre}. We then present our Sharing MultiHead-MoA(SMMoA) and block-specific design, especially Prompt Generator(PG) in Sec. \ref{sec:share-moa} and Sec. \ref{sec:block-specific}, respectively. Finally, we introduce the optimization function of the framework briefly in Sec. \ref{sec:loss}.

\subsection{Preliminaries}\label{sec:Pre}
\textbf{Adapters} are implemented by embedding compact, auxiliary layers into the structure of Transformer architecture. Typically, an adapter layer consists of a down-projection matrix $W_{\mathrm{down}} \in \mathbb{R}^{r \times d}$, followed by a non-linear activation function $\sigma(\cdot)$, and an up-projection matrix $W_{\mathrm{up}} \in \mathbb{R}^{d \times r}$. In this context, \emph{d} represents the dimension of the hidden layer, and \emph{r} serves as the bottleneck dimension, which is a hyperparameter used in configuring the adapters. Denote \emph{$x$} as the input to the adapter, the computation within the adapter module (with residual) can be summarized as follows:
\begin{equation}
\label{equation:standard adapter} 
\operatorname{Adapter}(x)=W_{\text {up }} \sigma\left(W_{\text {down }} x\right)+x
\end{equation}

\textbf{Mixture-of-Experts~(MoE)}~\citep{moec} is a promising approach to scale up the number of parameters within the same computational bounds.
Different from standard transformer models, each MoE layer consists of $N$ independent feed-forward networks $\{\boldsymbol{E}_i\}^{N}_{i=0}$ as the experts, along with a gating function $\alpha \left(\cdot\right)$ to model a probability distribution indicating the weights over these experts' outputs.
For the hidden representation $\boldsymbol{h} \in \mathbb{R}^{d}$ of input token, the gate value of routing $\boldsymbol{h}$ to expert $\boldsymbol{E}_i$ is denoted as:
\begin{equation}
\alpha \left(\boldsymbol{E}_i\right) = \exp\left(\boldsymbol{h}\cdot\boldsymbol{e}_i\right) / \sum_{j=0}^{N}\exp\left(\boldsymbol{h}\cdot\boldsymbol{e}_j\right),
\end{equation}
where $\boldsymbol{e}_i$ denotes the trainable embedding of $\boldsymbol{E}_i$. Then, the corresponding $k$ experts, according to the top-$k$ gated values, are activated and the output $\boldsymbol{O}$ of the MoE layer is
\begin{equation}
\boldsymbol{O} = \boldsymbol{h} + \sum_{i=0}^{N}\alpha \left(\boldsymbol{E}_i\right) \cdot \boldsymbol{E}_i\left(\boldsymbol{h}\right).
\end{equation}

\subsection{Sharing MoA}\label{sec:share-moa}

\textbf{Sharing Strategy.} 
As shown in Fig. \ref{fig:AdapterX_framework},  
\methodshort{} adopts parameter sharing across transformer blocks to consider the allocation of trainable parameters from an inter-layer perspective. 
Instead of sharing the only same adapter across all layers like \cite{kopiczko2023vera}, \cite{renduchintala2023tied}, Mixture of Adapters(MoA) is used to improve model flexibility since the router can select the most suitable adapter expert for each block. With this, we achieve the parameter allocation rather than simply sharing.
In order to allow routers and experts to learn the features of all blocks for more accurate and comprehensive parameter allocation, 
\methodshort{} uses the same router and adapter expert repository across all blocks. Formally, let $X^1=\{x_1^1,x_2^1,\dots,x_L^1\}$  represent the input hidden representations of the first Transformer block. We define parameter sharing within our model as $X^{i+1}=\mathrm{MoA}(X^i)$, which deviates from the approach used in existing Mixture of Experts (MoE) models, where $X^{i+1}=\mathrm{MoE}^i(X^i)$ is typical.

It is important to note a key distinction in our approach: while the MoE layer, including the router, shares trainable parameters across the model, the representations for the same token are not identical in every transformer block. This means that the representation of a given token, denoted as $h_i^j$, at layer $j$, may be directed to a different expert compared to its representation, $h_i^j+1$, at layer $j+1$. 
Consequently, on blocks in different locations, even for the same token, due to its different features and block locations, the router is still very likely to assign different adapter experts to it. 
Based on this principle, we implement token-level allocation. And this approach is validated in Sec.~\ref{sec:visualization}.

\textbf{Detailed design of MoA.} To handle features in more detail, we first split the input hidden states into multi-heads, then conduct a projection over the sub hidden states to the expert embedding space with lower dimension. Formally, we denote a sequence of inputs tokens by  $\mathbf{X} \in \mathbb{R}^{l \times d}$, where $l$ is the number of tokens and $d$ represents the length of token dimension. First, every token in  $X$ is split into $h$ sub-tokens along the token dimensions, and these subtokens are arranged in parallel according to the original token sequence, forming a new feature space $\hat{\mathbf{X}} \in \mathbb{R}^{(l\times h) \times \frac{d}{h}}$ as:
\begin{align}
    \hat{\mathbf{X}} &=  \digamma_{\text{s}} ({\mathbf{X}}) \nonumber \\
    &= \left[\underbrace{\overbrace{\mathbf{x}^{0}_0,\ldots, \mathbf{x}^{0}_{h-1}}^{h}, \ldots ,\mathbf{x}^{i}_{j},\mathbf{x}^{i}_{j+1},\ldots,\mathbf{x}^{l}_{h-1}}_{l \times h}\right],
\label{eq: sub-tokens}
\end{align}

where function $\digamma_{\text{s}}$ denotes the token splitting operation: $\mathbb{R}^{l \times d} \rightarrow \mathbb{R}^{\left(l \times h\right) \times \frac{d}{h}}$, and each sub-token is presented as $\mathbf{x}^{i}_{j} \in \mathbb{R}^{\frac{d}{h}}$, meaning it is the the $j^{th}$ sub-token split from the $i^{th}$ token.

Then, we first parameterize the experts with lower-dimensional embeddings  $\mathbf{e}^{i} \in \mathbb{R}^{d_e}$ such that $d_e$ is much smaller than the Transformer hidden size $d$, to be simple, $d_e$ is defined same as the rank of adapters. And we normalize it with a frozen $L_2$ norm function to get the result $\mathbf{e}^{in}_{norm}$. For input token $\hat{\mathbf{X}}$, we use a linear projection $f^{proj}(\hat{\mathbf{X}})$ such that ${\mathbf{W}} \in \mathbb{R}^{d_e \times d}$. Thus, the routing scoring function between the tokens and experts can be written as $s_i = ({\mathbf{W}}\hat{\mathbf{X}}) \cdot \mathbf{e}^{in}_{norm}$. 
The gating value of routing $\mathbf{h}$ to expert $f^{\text{FFN}}_{i}$ is denoted as:
\begin{equation}
g \left(f^{\text{FFN}}_{i}\right) = \exp\left(s_i\right) / \sum_{j=0}^{N}\exp\left(s_i\right),
\label{Eq: Softmax in Routing}
\end{equation}
With the multi-head split and lower-dimension projector, the dimension of input can gradually decrease a lot, which fits the number of experts better than the larger hidden size of Transformer. As a result, the combination of these two design can better fits in with the low-rank nature of routing.
The algorithm flow is shown in Algorithm~\ref{algo:algo}. 

\subsection{Block-Specific Design}\label{sec:block-specific}
As outlined in Sec.~\ref{sec:introduction}, 
our methodology leverages the strengths of its predecessor. While embracing a block-general perspective, we do not entirely discard block-specific design principles. In fact, we highlight the importance of providing diverse input representations for each block, which serves to improve the model's adaptability to a variety of datasets. To achieve this diversity, we have implemented two key design strategies.

\textbf{Different norm layer for each block.} We keep the norm layer before FFN different across blocks. Given that layer normalization operates with just two small vectors, introducing individual normalization layers adds only a minimal increase in trainable parameters to our framework.
By employing separate LayerNorm for each block, rather than a shared one, we introduce this separate normalization scheme that promotes a more varied representation of inputs. This tailored approach allows each block to capture a broader range of features.

\textbf{Different generated prompt for each block.} To enhance the distinctiveness of inputs fed into each transformer block, we propose a Prompt Generator (PG) block to leverage the output from the MultiHead-MoA layer, as illustrated in Fig. \ref{fig:AdapterX_framework}. The process begins with reducing the number of input tokens through an averaging mechanism. Subsequently, we introduce a scale and shift operation on the averaged representation to generate a unique prompt for each block. This prompt is then concatenated with output tokens. This simple yet effective technique cleverly combines the strengths of MultiHead-MoA and prompt tuning methods, thereby enriching the diversity of inputs across different transformer blocks.

\subsection{Objective Function}\label{sec:loss}
Following~\citep{fedus2022switch}, given the sub-token set $\hat{\mathbf{X}}$ (depicted in Eq.~\ref{eq: sub-tokens}) and the frequency $t_p$ of how many sub-tokens are routed to the $p^{th}$ expert, we compute the load balancing loss $\mathcal{L}_{\text{balance}}$ via:
\begin{equation}
 \mathcal{L}_{\text{balance}} = \frac{N}{|\hat{\mathbf{X}}|} \sum_{p=1}^{N}\sum_{\mathbf{x}^{i}_{j} \in \hat{\mathbf{X}}} t_p \cdot g \left(f^{\text{FFN}}_{p}\right),
\end{equation}
where $N$ denotes the number of experts, $|\hat{\mathbf{X}}|$ is the number of sub-tokens contained in $\hat{\mathbf{X}}$. $g \left(f^{\text{FFN}}_{p}\right)$ is the gating function depicted in Eq.~\ref{Eq: Softmax in Routing}, denoting the gating value of routing a certain sub-token $\mathbf{x}^{i}_{j}$ into the $p^{th}$ expert.
The overall training objective is to minimize:
\begin{equation}
\label{eq:loss}
    \mathcal{L} = \mathcal{L}_\text{task} + \alpha \mathcal{L}_{\text{balance}},
\end{equation}
where $\alpha$ is a coefficient for load balancing, which is usually set to 0.01 in this work.

\section{Experiments}
\label{sec:exp}

We evaluate the effectiveness of \methodshort{} by conducting extensive visual recognition experiments in both the 2d image and 3d point cloud domains. 
For clarity, we define the adapter in \methodshort{} as a modular component. This definition facilitates seamless integration with other methods. For instance, when integrated with Adaptformer, the resulting approach is called Adaptformer-X.

We first describe our experimental settings in Sec. \ref{Sec:exp_setting}, covering pre-trained backbones, baseline methods, downstream tasks and training details. 
We then compare with baseline methods and provide a thorough analysis in Sec. \ref{Sec:exp_2d} and Sec. \ref{Sec:exp_3d}. In addition, we also conduct ablation studies to explore different experimental configurations and explain what makes for the superiority of \methodshort{} in Sec \ref{sec:exp_ab} and visual experiments in Sec \ref{sec:visualization} for better understanding.

\subsection{Experimental Settings}\label{Sec:exp_setting}

\textbf{Pre-trained backbone.}
For \textcolor{Image}{\textbf{image}} modality, we adopt the plain Vision Transformer~(ViT)~\cite{dosovitskiy-2020-vit}, \ie,ViT-Base~(ViT-B/16) as our backbone model, and directly use the ImageNet-21k~\cite{deng-cvpr09-imagenet} supervised pre-trained model.
For \textcolor{Point Cloud}{\textbf{point cloud}} modality, we take widely used pre-trained models from PointMAE~\cite{pang2022masked}. More details about pre-training approaches and datasets can be found in Appendix.

\textbf{Baseline methods.} 
In \textcolor{Image}{\textbf{image}} domain, we compare \methodshort{} with (1)\emph{Standard Adapter}~\cite{houlsby2019parameter}: adding the adapter after the FFN module in serialeze, and (2) \emph{AdaptFormer}~\cite{chen2022adaptformer}: adding adapter after the FFN module in parallel. 
For \textcolor{Point Cloud}{\textbf{point cloud}}, we compare \methodshort{} with \emph{DAPT}~\cite{zhou2024dapt}: generating a unique scale for each token to dynamically adjust features by considering the significance score.

\textbf{Downstream tasks.}
We evaluate our \methodshort{} on both image and point cloud recognition tasks to verify its effectiveness. The specific datasets leveraged in this work are presented in the following.

\textcolor{Image}{\textbf{$\bullet$\quad Image domain :}}
VTAB-1K~\cite{vtab} contains 19 image classification tasks from diverse fields, which can be categorized into three groups: Natural, Specialized, and Structured. These tasks cover a large range of possible domains where downstream tasks come, so the performance of different methods on this benchmark largely reflects their ability to transfer learning. Each dataset contains 800 samples for training and 200 for validation. 

\textcolor{Point Cloud}{\textbf{$\bullet$\quad Point Cloud domain :}}
Tasks in point cloud classification can be divided into two categories, namely Real-World Object Classification and Synthetic Object Classification. The former is represented by ScanObjectNN~\cite{uy2019revisiting}, which consists of $\sim$$15$K point clouds of indoor objects obtained by scanning across $15$ categories. This benchmark is challenging since objects contain points from cluttered backgrounds and occlusion caused by other objects.
As shown in Tab.~\ref{tab:3dclass}, we conducted experiments on three variants of ScanObjectNN (\ie, OBJ\_BG, OBJ\_ONLY, and PB\_T50\_RS), each with increasing complexity.
As a representative of the synthesis task, ModelNet40~\cite{wu20153d} contains $12,311$ 3D CAD models across $40$ categories. As a synthetic dataset, it consists of complete, uniform, and noise-free point clouds, which is totally different from ScanObjectNN.

\textbf{Implement Details.}
To ensure a fair comparison, we employed identical experimental settings to the default fine-tuning method for each baseline. This entails freezing the weights of the pre-trained point cloud backbone and solely updating the newly inserted parameters during training. All experiments are conducted on a single NVIDIA A100 GPU.

\begin{table*}[t]
\caption{\textbf{Full results on the VTAB-1K benchmark}. ``Avg. Acc.'' denotes the average results over three groups. ``Params'' denotes the number of learnable parameters excluding the final classification layer, as the number of parameters in this final layer depend on the number of classes, which varies between 2 and 397. }
\vspace{-0.3cm}
\begin{center}
\setlength{\tabcolsep}{0.3pt}
\resizebox{\textwidth}{!}{
\begin{tabular}{p{3cm}<{}p{1.25cm}<{\centering}p{0.75cm}<{\centering}|p{0.75cm}<{\centering}p{0.75cm}<{\centering}p{0.75cm}<{\centering}p{0.75cm}<{\centering}p{0.75cm}<{\centering}p{0.75cm}<{\centering}p{0.75cm}<{\centering}|p{0.75cm}<{\centering}p{0.75cm}<{\centering}p{0.75cm}<{\centering}p{0.75cm}<{\centering}|p{0.75cm}<{\centering}p{0.75cm}<{\centering}p{0.75cm}<{\centering}p{0.75cm}<{\centering}p{0.75cm}<{\centering}p{0.75cm}<{\centering}p{0.75cm}<{\centering}p{0.75cm}<{\centering}}
\toprule
\multicolumn{3}{c|}{}&\multicolumn{7}{c|}{\textbf{Natural}}&\multicolumn{4}{c|}{\textbf{Specialized}}&\multicolumn{8}{c}{\textbf{Structured}}\\
&\multicolumn{1}{c}{\STAB{\rotatebox[origin=c]{90}{Params. (M)}}}
&\multicolumn{1}{c|}{\STAB{\rotatebox[origin=c]{90}{Avg. Acc.}}}
&\multicolumn{1}{c}{\STAB{\rotatebox[origin=c]{90}{Cifar100}}}
&\multicolumn{1}{c}{\STAB{\rotatebox[origin=c]{90}{Caltech101}}}
&\multicolumn{1}{c}{\STAB{\rotatebox[origin=c]{90}{DTD}}}
&\multicolumn{1}{c}{\STAB{\rotatebox[origin=c]{90}{Flower102}}}
&\multicolumn{1}{c}{\STAB{\rotatebox[origin=c]{90}{Pets}}}
&\multicolumn{1}{c}{\STAB{\rotatebox[origin=c]{90}{SVHN}}}
&\multicolumn{1}{c|}{\STAB{\rotatebox[origin=c]{90}{Sun397}}}
&\multicolumn{1}{c}{\STAB{\rotatebox[origin=c]{90}{Camelyon}}}
&\multicolumn{1}{c}{\STAB{\rotatebox[origin=c]{90}{EuroSAT}}}
&\multicolumn{1}{c}{\STAB{\rotatebox[origin=c]{90}{Resisc45}}}
&\multicolumn{1}{c|}{\STAB{\rotatebox[origin=c]{90}{Retinopathy}}}
&\multicolumn{1}{c}{\STAB{\rotatebox[origin=c]{90}{Clevr-Count}}}
&\multicolumn{1}{c}{\STAB{\rotatebox[origin=c]{90}{Clevr-Dist}}}
&\multicolumn{1}{c}{\STAB{\rotatebox[origin=c]{90}{DMLab}}}
&\multicolumn{1}{c}{\STAB{\rotatebox[origin=c]{90}{KITTI-Dist}}}
&\multicolumn{1}{c}{\STAB{\rotatebox[origin=c]{90}{dSpr-Loc}}}
&\multicolumn{1}{c}{\STAB{\rotatebox[origin=c]{90}{dSpr-Ori}}}
&\multicolumn{1}{c}{\STAB{\rotatebox[origin=c]{90}{sNORB-Azim}}}
&\multicolumn{1}{c}{\STAB{\rotatebox[origin=c]{90}{sNORB-Ele}}}\\
\specialrule{0em}{1pt}{1pt}
\hline
\specialrule{0em}{1pt}{1pt}
\multicolumn{22}{l}{\emph{Conventional Fine-Tuning}}\\
\specialrule{0em}{1pt}{1pt}
\sc Full&85.84&68.9&68.9&87.7&64.3&97.2&86.9&87.4&38.8&79.7&95.7&84.2&73.9&56.3&58.6&41.7&65.5&57.5&46.7&25.7&29.1 \\
\sc Linear&0.04&57.6&64.4&85.0&63.2&97.0&86.3&36.6&51.0&78.5&87.5&68.5&74.0&34.3&30.6&33.2&55.4&12.5&20.0&9.6&19.2\\
\hline
\specialrule{0em}{1pt}{1pt}
\multicolumn{22}{l}{\emph{Vision PEFT methods}}\\
\specialrule{0em}{1pt}{1pt}
{\sc VPT-Deep}~\cite{vpt}&0.53&72.0&78.8&90.8&65.8&98.0&88.3&78.1&49.6&81.8&96.1&83.4&68.4&68.5&60.0&46.5&72.8&73.6&47.9&32.9&37.8 \\
{\sc NOAH}~\cite{noah}&0.36&75.5&69.6&92.7&70.2&99.1&90.4&86.1&53.7&84.4&95.4&83.9&75.8&82.8&68.9&49.9&81.7&81.8&48.3&32.8&44.2\\
{\sc LoRA}\cite{lora}&0.29&74.5&67.1&91.4&69.4&98.8&90.4&85.3&54.0&84.9&95.3&84.4&73.6&82.9&69.2&49.8&78.5&75.7&47.1&31.0&44.0\\
{\sc SSF}~\cite{ssf}&0.24&75.7&69.0&92.6&75.1&99.4&91.8&90.2&52.9&87.4&95.9&87.4&75.5&75.9&62.3&53.3&80.6&77.3&54.9&29.5&37.9\\
{\sc BitFit}~\cite{bitfit}&0.10&65.2&72.8&87.0&59.2&97.5&85.3&59.9&51.4&78.7&91.6&72.9&69.8&61.5&55.6&32.4&55.9&66.6&40.0&15.7&25.1\\
\hline
\specialrule{0em}{1pt}{1pt}
\multicolumn{22}{l}{{\sc {Adapter}\cite{houlsby2019parameter}} (Baseline)}\\
\specialrule{0em}{1pt}{1pt}
\quad\ $rank=8$&{0.15}&73.9&70.6&91.7&68.2&98.9&90.5&82.7&54.1&82.9&95.3&82.2&75.9&81.6&65.3&48.1&76.2&76.2&47.4&29.3&41.0\\
\quad\ $rank=64$&{1.19}&74.2&68.5&92.3&67.8&98.8&90.2&85.1&53.2&84.2&95.9&83.9&75.3&80.9&63.7&49.1&80.2&75.9&48.9&31.5&36.1\\
\rowcolor{linecolor!40}\multicolumn{22}{l}{{\sc {Adapter-X}} (Ours)}\\
\rowcolor{linecolor!40}\quad\ $rank=64$&\bf {0.17}&\bf{74.3}&69.5&92.7&70.9&99.2&89.9&85.4&53.3&85.5&95.9&84.8&74.0&82.2&64.0&48.0&78.5&75.8&44.8&29.0&40.1\\

\hline
\specialrule{0em}{1pt}{1pt}
\multicolumn{22}{l}{{\sc {AdaptFormer}\cite{chen2022adaptformer}} (Baseline)}\\
\specialrule{0em}{1pt}{1pt}
\quad\ $rank=64$&{1.26}&76.1&73.1&92.6&71.2&99.2&91.3&89.7&55.2&86.9&95.4&86.0&75.0&82.5&62.3&51.9&78.9&86.4&52.7&33.9&36.5\\
\rowcolor{linecolor!40}\multicolumn{22}{l}{{\sc {AdaptFormer-X}} (Ours)}\\
\rowcolor{linecolor!40}\quad\ $rank=64$&\bf {0.17}&\bf{76.2}&73.4&93.3&71.8&99.3&91.4&87.8&56.7&87.5&95.5&84.8&75.8&80.7&62.0&50.5&79.0&85.8&53.7&32.7&41.2\\
\bottomrule
\end{tabular}
}
\end{center}
\label{tab:vtab}
\vspace{-0.5cm}
\end{table*}

\subsection{Experiment on Image}\label{Sec:exp_2d}
Tab.~\ref{tab:vtab} shows the comparison of our method with full-tuning, linear probing(\ie, only training the classification head), some representative vision PEFT methods and particular vision adapter methods. We conduct the expertiments based on two baselines: NOAH and AdaptFormer. The Former Adapter is serialized while the latter is parallel with scale. 

Our comparative experiments on the NOAH framework demonstrate that when our method's hyperparameters are aligned with those of the baseline, specifically the rank of the adapter, our approach achieves a reduction in the number of parameters by a factor of 7.4, while sustaining marginally superior performance. Furthermore, significant enhancements in performance are observed when our method is compared to the baseline with an equivalent number of parameters.
The same phenomenon can still be observed when experimenting with another widely used Adaptformer framework. 
This also shows that our method has relatively good applicability to different adapters.

\begin{table*}[ht]
    \footnotesize
    \centering
  \caption{
Classification on three variants of ScanObjectNN~\cite{uy2019revisiting} and ModelNet40~\cite{wu20153d} datasets, including the number of trainable parameters and overall accuracy (OA). Point cloud rotation is used for ScanObjectNN and scale\&transform is used for ModelNet40 following previous works, and $^\dag$ means the model is reproduced using the same fine-tuning configuration as ~\cite{dong2022autoencoders}. We report ScanObjectNN results without voting, ModelNet40 results without and with voting, referred to (-/-).
}
\label{tab:3dclass} 
  \resizebox{\textwidth}{!}{
    \begin{tabular}{lcccccccc}
    
    \toprule
    \multirow{2.3}{*}{Method} &\multirow{2.3}{*}{\parbox{2cm}{\centering Tunable params.(M)}} &\multicolumn{3}{c}{ScanObjectNN} &\multicolumn{2}{c}{ModelNet40}\\
		\cmidrule(r){3-5} \cmidrule{6-7}
	& &OBJ\_BG & OBJ\_ONLY &PB\_T50\_RS & Points Num. & OA (\%)      \\
    \midrule
    \multicolumn{7}{c}{\textit{Supervised Learning Only}} \\
    \midrule
    PointNet~\cite{qi2017pointnet} & 3.5 & 73.3  & 79.2  & 68.0 & 1k & - / 89.2 \\
    PointNet++~\cite{qi2017pointnet++}   & 1.5 & 82.3  & 84.3  & 77.9 & 1k & - / 90.7\\
    DGCNN~\cite{wang2019dynamic}  & 1.8 & 82.8  & 86.2  & 78.1 & 1k & - / 92.9 \\
    PointNeXt~\cite{qian2022pointnext}  & 1.4  & -     & -  & 87.7 & 1k & - / 94.0\\
    PointMLP~\cite{ma2022rethinking}  &  13.2 & -    & -     & 85.4  & 1k & - / 94.5\\
    \midrule
    \multicolumn{7}{c}{\textit{ Self-Supervised Representation Learning (Full fine-tuning)}} \\
    \midrule
    OcCo~\cite{wang2021unsupervised} & 22.1 & 84.85 & 85.54 & 78.79 & 1k & - / 92.1 \\
    Point-BERT~\cite{yu2022point}  & 22.1 & 87.43 & 88.12 &  83.07 & 1k & 92.7 / 93.2 \\
    MaskPoint~\cite{liu2022masked} & 22.1 & 89.70 & 89.30 &  84.60 & 1k & - / 93.8 \\
    Point-MAE~\cite{pang2022masked}  & 22.1 & 90.02 & 88.29 & 85.18 & 1k & 93.2 / 93.8 \\
    Point-M2AE~\cite{zhang2022point}  & 15.3 & 91.22 & 88.81 & 86.43 
 & 1k & 93.4 / 94.0\\
    ACT~\cite{dong2022autoencoders}   & 22.1 & 93.29 & 91.91  & 88.21 & 1k & - / 93.7\\
    Recon~\cite{qi2023recon} & 43.6 & 95.18 & 93.12  & 89.73 & 1k & 93.6 / 93.9  \\
    
    \midrule
    \multicolumn{7}{c}{\textit{Self-Supervised Representation Learning (Efficient fine-tuning)}} \\
    
    \midrule
    
    Point-MAE~\cite{pang2022masked}$^\dag$& 22.1 (100\%)& 91.74 & 90.88 & {88.31} & 1k & 93.2 / {\color{gray}{93.8}}\\
    + DAPT~\cite{zhou2024dapt} & 1.1 (4.97\%) & {92.08}\dplus{+0.34} & {91.22}\dplus{+0.34} & {87.13}\dtplus{\bf{-1.18}} & 1k & {93.5}{\dplus{+0.3}} / {\color{gray}{{94.0}}}{\color{gray}{\ddplus{+0.2}}} \\
    
    \rowcolor{linecolor!40}+ DAPT-X\textbf{(ours)} & \textbf{0.42} (\textbf{1.88}\%) & {92.60}\dplus{+0.96} & {92.43}\dplus{+1.55} & {88.45}\dplus{\bf{+0.14}} & 1k & {93.5}{\dplus{+0.3}} / {\color{gray}{{94.1}}}{\color{gray}{\ddplus{+0.3}}} \\
    \bottomrule
    \end{tabular}
}

\end{table*}

\subsection{Experiment on Point Cloud}\label{Sec:exp_3d}
The experimental results of our \methodshort{} on Point Cloud classification tasks are shown in Tab.~\ref{tab:3dclass}. Point-MAE~\cite{pang2022masked} are adopted as our baseline. We take the SOTA point cloud efficient fine-tuning method DAPT~\cite{zhou2024dapt} into comparison and report results on ScanObjectNN and ModelNet40. Experimental results presented in the last three rows of the table indicates that our \methodshort{} meets or exceeds the performance of full fine-tuning and DAPT in all cases with less tunable parameters. Specially, \methodshort{} achieves accuracy increase of $1.55\%$ over Point-MAE on OBJ\_ONLY and $1.32\%$ over DAPT on PB\_T50\_RS with only less than half of tunable parameters of DAPT. The results demonstrates the effectiveness and generalization ability of the mixture of adapters and block-specific operations in our \methodshort{}.

\subsection{Ablation Studies}\label{sec:exp_ab}

\begin{figure*}
\centering
\includegraphics[width=0.7\linewidth]{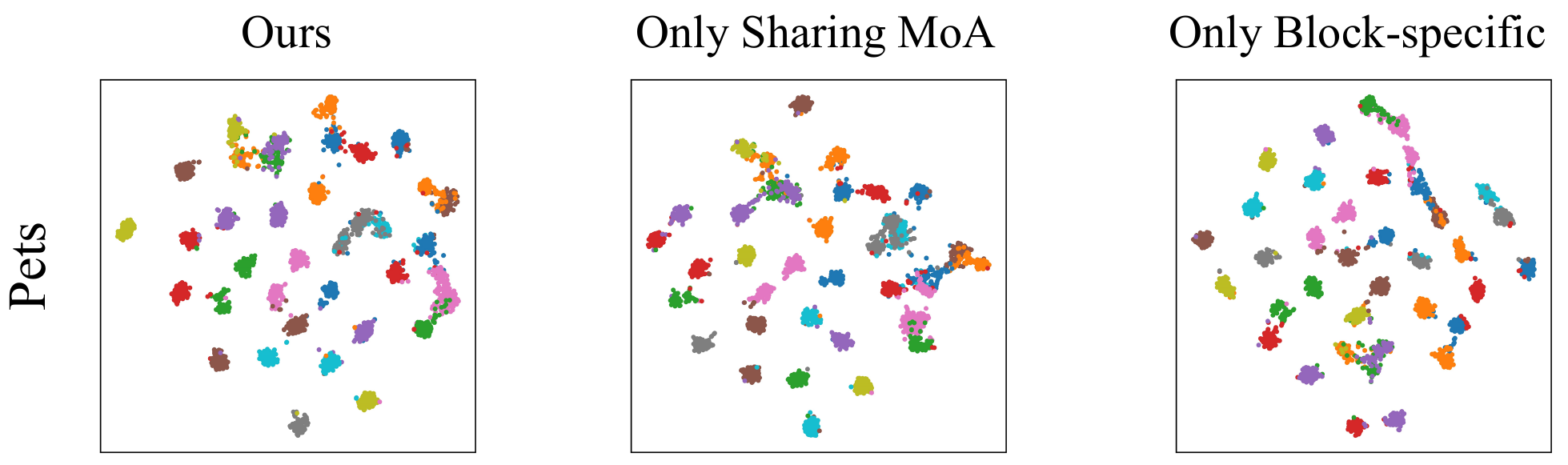}
\caption{Partial t-SNE visualization results of our ablation study on the 2D VTAB-1K dataset. Please refer to the appendix for more results.}
\label{fig:visualize_tsne}
\vspace{-0.5cm}
\end{figure*}

We conducted ablation studies on \methodshort{} to validate our core design assumptions stated in Sec.~\ref{sec:introduction}, with 2D experiments on VATB and 3D on ScanObjNN-PB\_T50\_RS datasets.

The 2D ablation reveals that using only the Sharing Module of Adapters (SMoA) in \methodshort{} reduces parameters, linked to the specific second norm layer and Prompt Generator in each block, which is little. But removing these leads to decreased adaptability. However, eliminating sharing increases parameters and improves performance, consistent with the reasons outlined in Sec.~\ref{sec:introduction}. Compared to methods relying heavily on block-specific parameters, our approach has over ten times fewer parameters with only little performance drop, indicating a balanced efficiency and performance.

In the domain of 3D PEFT, the performance improvement brought by the larger number of parameters does not always translate to benefits due to the susceptibility of point cloud datasets to overfitting, as shown in Tab.\ref{tab:ablation on 2&3d}(b). Although more parameters enhance performance, considering all four advantages, \methodshort{} achieves significant performance with fewer parameters.

\begin{table*}[t]
\caption{Effect of different perspective, namely block-general and block-specific. Ablation experiments are conducted using the VTAB dataset on the 2d  \textcolor{Image}{\textbf{image}} modality and the xxx dataset on the 3d  \textcolor{Point Cloud}{\textbf{point cloud}} modality.
}
\label{tab:ablation on 2&3d}
    \centering
 \begin{minipage}{0.49\linewidth}
  \centering
        \subcaption{2d test accuracy}
\resizebox{!}{0.9cm}
        {
\begin{tabular}{ccc}
\hline
\begin{tabular}[c]{@{}c@{}}Different\\ design\end{tabular} &
  \begin{tabular}[c]{@{}c@{}} Acc.\end{tabular} &
  \begin{tabular}[c]{@{}c@{}} Total \\ Params.(M)\end{tabular} \\ \hline
Both  & 76.2 & 0.17 \\
Only Sharing MoA  & 74.9 & 0.15       \\
Only Block-specific & 76.4 & 1.8       \\ \hline
\end{tabular}
}
 \end{minipage}
 \hfill
 \begin{minipage}{0.49\linewidth}
  \centering
        \subcaption{3d test accuracy}
\resizebox{!}{0.9cm}
        {
\begin{tabular}{ccc}
\hline
\begin{tabular}[c]{@{}c@{}}Different\\ design\end{tabular} &
  \begin{tabular}[c]{@{}c@{}} Acc.\end{tabular} &
  \begin{tabular}[c]{@{}c@{}} Total \\ Params.\end{tabular} \\ \hline
Both  & 88.45 & 0.56 \\
Only Sharing MoA  & 87.82 & 0.46         \\
Only Block-specific & 87.99 & 1.40         \\ \hline
\end{tabular}
}
\end{minipage}
\vspace{-2mm}
\end{table*}

\subsection{Visualization}\label{sec:visualization}
\begin{figure*}
\centering
\includegraphics[width=0.93\linewidth]{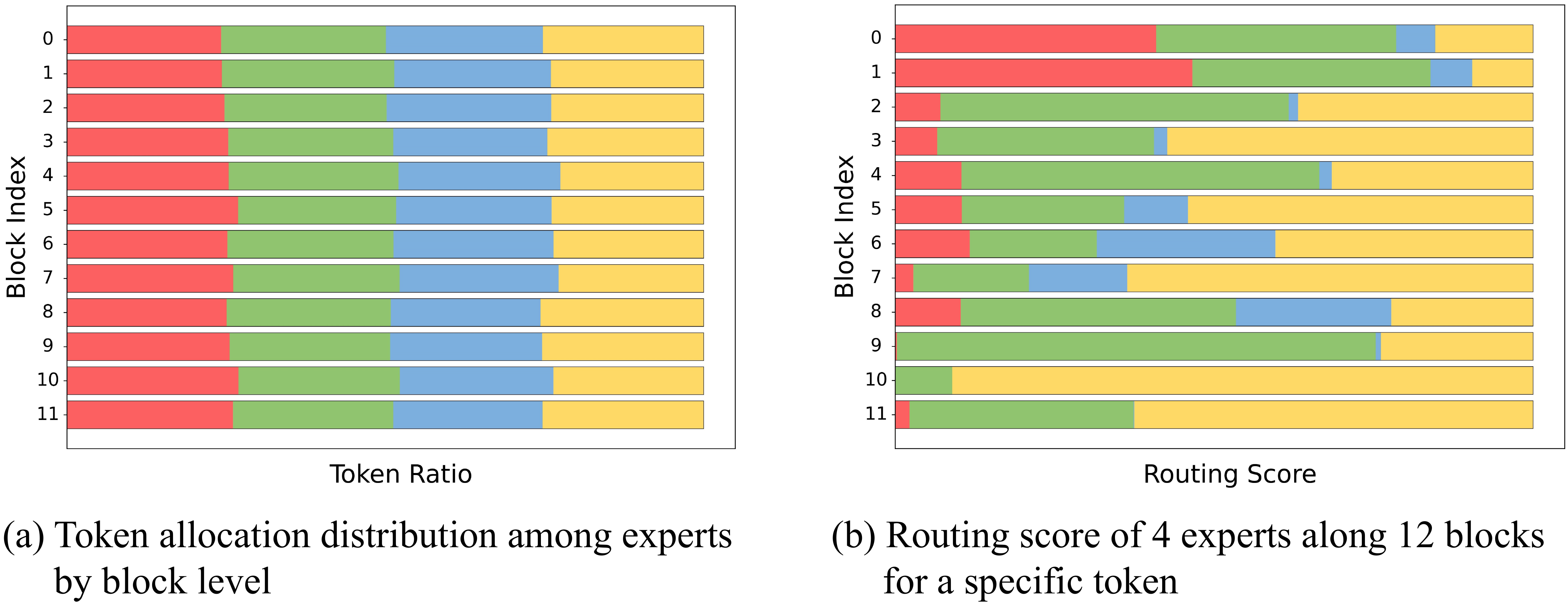}
\caption{The distribution of token allocation and routing score of 4 shared experts in different blocks on the CIFAR100 dataset of VTAB-1K.}
\label{fig:visualize_expert}
\vspace{-0.5cm}
\end{figure*}

To further illustrate how our method achieves surprising results,
we conduct a two-part visualization experiment. Fig.~\ref{fig:visualize_tsne} is conducted on Pets validation set from VTAB-1K dataset, and Fig.\ref{fig:visualize_expert} is conducted on the CIFAR100 dataset of VTAB-1K.

In order to better evaluate the necessity of model design, we further conduct t-SNE~\cite{van2008visualizing} visualizations on the complete \methodshort{} and only retain part of the design based on the ablation experiments. The features are harvested from the Pets validation subset of the VTAB-1K dataset, processed through the ViT-Base backbone. Fig.~\ref{fig:visualize_tsne} shows that  both the SMoA alone and the block-specific modules alone result in somewhat blended feature outputs. Compared with the above two methods, the full strategy performs well in projecting features.

Secondly, as shown in Fig.~\ref{fig:visualize_expert}, the left picture represents the ratio of the number of tokens accepted by four experts at different block positions for the whole dataset. The right figure refers to the probability that a randomly selected token is assigned to four different experts in different blocks during its forward progress. Different colors are used in both figures to represent the strategies of different experts.
The picture on the left shows that at the dataset level, adapter experts are selected with roughly equivalent probability for every block, thereby implicitly increasing the parameter count that can be trained per block. The right figure shows that the identical token will be allocated to different experts in different blocks for processing, substantiating that \methodshort{} can indeed achieve token-level dynamics.

\vspace{-0.15cm}
\section{Conclusion}
\vspace{-0.15cm}
Our analysis of key in PEFT has highlighted the importance of parameter sharing for redundancy reduction and the role of tunable parameters, dynamic allocation, and block-specific design in enhancing performance. Inspired by these findings, we have introduced Adapter-X, featuring a Sharing Mixture of Adapters (Sharing MoA) module that achieves token-level dynamic allocation, implicitly increases tunable parameters per block, and facilitates inter-block sharing. The incorporation of block-specific designs like the Prompt Generator (PG) further refines the model's adaptability.
Extensive experiments on 2D images and 3D point clouds have shown that Adapter-X is a significant advancement, being the first to exceed the performance of full fine-tuning with a substantially lower parameter count.

{\small
\bibliographystyle{abbrv}
\bibliography{main}

\begin{thebibliography}{10}

\bibitem{chen2023parameter}
J.~Chen, A.~Zhang, X.~Shi, M.~Li, A.~Smola, and D.~Yang.
\newblock Parameter-efficient fine-tuning design spaces.
\newblock {\em arXiv preprint arXiv:2301.01821}, 2023.

\bibitem{chen2022adaptformer}
S.~Chen, C.~Ge, Z.~Tong, J.~Wang, Y.~Song, J.~Wang, and P.~Luo.
\newblock Adaptformer: Adapting vision transformers for scalable visual recognition.
\newblock {\em Advances in Neural Information Processing Systems}, 35:16664--16678, 2022.

\bibitem{chi2022representation}
Z.~Chi, L.~Dong, S.~Huang, D.~Dai, S.~Ma, B.~Patra, S.~Singhal, P.~Bajaj, X.~Song, X.-L. Mao, et~al.
\newblock On the representation collapse of sparse mixture of experts.
\newblock {\em Advances in Neural Information Processing Systems}, 35:34600--34613, 2022.

\bibitem{Dehghani2018}
M.~Dehghani, S.~Gouws, O.~Vinyals, J.~Uszkoreit, and {\L}.~Kaiser.
\newblock Universal transformers.
\newblock {\em arXiv preprint arXiv:1807.03819}, 2018.

\bibitem{deng-cvpr09-imagenet}
J.~Deng, W.~Dong, R.~Socher, L.-J. Li, K.~Li, and L.~Fei-Fei.
\newblock Imagenet: A large-scale hierarchical image database.
\newblock In {\em IEEE/CVF Conference on Computer Vision and Pattern Recognition}, 2009.

\bibitem{dong2022autoencoders}
R.~Dong, Z.~Qi, L.~Zhang, J.~Zhang, J.~Sun, Z.~Ge, L.~Yi, and K.~Ma.
\newblock Autoencoders as cross-modal teachers: Can pretrained 2d image transformers help 3d representation learning?
\newblock In {\em ICLR}, 2022.

\bibitem{dosovitskiy-2020-vit}
A.~Dosovitskiy, L.~Beyer, A.~Kolesnikov, D.~Weissenborn, X.~Zhai, T.~Unterthiner, M.~Dehghani, M.~Minderer, G.~Heigold, S.~Gelly, et~al.
\newblock An image is worth 16x16 words: Transformers for image recognition at scale.
\newblock {\em arXiv preprint arXiv:2010.11929}, 2020.

\bibitem{edalati2022krona}
A.~Edalati, M.~Tahaei, I.~Kobyzev, V.~P. Nia, J.~J. Clark, and M.~Rezagholizadeh.
\newblock Krona: Parameter efficient tuning with kronecker adapter.
\newblock {\em arXiv preprint arXiv:2212.10650}, 2022.

\bibitem{fedus2022switch}
W.~Fedus, B.~Zoph, and N.~Shazeer.
\newblock Switch transformers: Scaling to trillion parameter models with simple and efficient sparsity.
\newblock {\em The Journal of Machine Learning Research}, 23(1):5232--5270, 2022.

\bibitem{Ge2022}
T.~Ge, S.-Q. Chen, and F.~Wei.
\newblock Edgeformer: A parameter-efficient transformer for on-device seq2seq generation.
\newblock {\em arXiv preprint arXiv:2202.07959}, 2022.

\bibitem{he2021towards}
J.~He, C.~Zhou, X.~Ma, T.~Berg-Kirkpatrick, and G.~Neubig.
\newblock Towards a unified view of parameter-efficient transfer learning.
\newblock In {\em Proc. of Intl. Conf. on Learning Representations}, 2021.

\bibitem{sparseadapter}
S.~He, L.~Ding, D.~Dong, J.~Zhang, and D.~Tao.
\newblock Sparseadapter: An easy approach for improving the parameter-efficiency of adapters.
\newblock In {\em Findings of EMNLP}, 2022.

\bibitem{pevit}
X.~He, C.~Li, P.~Zhang, J.~Yang, and X.~E. Wang.
\newblock Parameter-efficient model adaptation for vision transformers.
\newblock In {\em Proceedings of AAAI}, 2023.

\bibitem{houlsby2019parameter}
N.~Houlsby, A.~Giurgiu, S.~Jastrzebski, B.~Morrone, Q.~De~Laroussilhe, A.~Gesmundo, M.~Attariyan, and S.~Gelly.
\newblock Parameter-efficient transfer learning for nlp.
\newblock In {\em International conference on machine learning}, pages 2790--2799. PMLR, 2019.

\bibitem{lora}
E.~J. Hu, yelong shen, P.~Wallis, Z.~Allen-Zhu, Y.~Li, S.~Wang, L.~Wang, and W.~Chen.
\newblock Lo{RA}: Low-rank adaptation of large language models.
\newblock In {\em Proceedings of ICLR}, 2022.

\bibitem{hu2023llm}
Z.~Hu, L.~Wang, Y.~Lan, W.~Xu, E.-P. Lim, L.~Bing, X.~Xu, S.~Poria, and R.~K.-W. Lee.
\newblock Llm-adapters: An adapter family for parameter-efficient fine-tuning of large language models.
\newblock {\em arXiv preprint arXiv:2304.01933}, 2023.

\bibitem{vpt}
M.~Jia, L.~Tang, B.~Chen, C.~Cardie, S.~J. Belongie, B.~Hariharan, and S.~Lim.
\newblock Visual prompt tuning.
\newblock In {\em Proceedings of ECCV}, 2022.

\bibitem{jie2023revisiting}
S.~Jie, H.~Wang, and Z.-H. Deng.
\newblock Revisiting the parameter efficiency of adapters from the perspective of precision redundancy.
\newblock In {\em Porc. of IEEE Intl. Conf. on Computer Vision}, 2023.

\bibitem{kopiczko2023vera}
D.~J. Kopiczko, T.~Blankevoort, and Y.~M. Asano.
\newblock Vera: Vector-based random matrix adaptation.
\newblock {\em arXiv preprint arXiv:2310.11454}, 2023.

\bibitem{lei2024conditional}
T.~Lei, J.~Bai, S.~Brahma, J.~Ainslie, K.~Lee, Y.~Zhou, N.~Du, V.~Zhao, Y.~Wu, B.~Li, et~al.
\newblock Conditional adapters: Parameter-efficient transfer learning with fast inference.
\newblock {\em Advances in Neural Information Processing Systems}, 36, 2024.

\bibitem{li2021prefix}
X.~L. Li and P.~Liang.
\newblock Prefix-tuning: Optimizing continuous prompts for generation.
\newblock In {\em Annual Meeting of the Association for Computational Linguistics}, 2021.

\bibitem{ssf}
D.~Lian, D.~Zhou, J.~Feng, and X.~Wang.
\newblock Scaling {\&} shifting your features: {A} new baseline for efficient model tuning.
\newblock In {\em Proceedings of NeurIPS}, 2022.

\bibitem{liu2022masked}
H.~Liu, M.~Cai, and Y.~J. Lee.
\newblock Masked discrimination for self-supervised learning on point clouds.
\newblock In {\em ECCV}, 2022.

\bibitem{Lou2021}
Q.~Lou, T.~Hua, Y.-C. Hsu, Y.~Shen, and H.~Jin.
\newblock Dictformer: Tiny transformer with shared dictionary.
\newblock In {\em International Conference on Learning Representations}, 2021.

\bibitem{ma2022rethinking}
X.~Ma, C.~Qin, H.~You, H.~Ran, and Y.~Fu.
\newblock Rethinking network design and local geometry in point cloud: A simple residual mlp framework.
\newblock In {\em ICLR}, 2022.

\bibitem{compactor}
R.~K. Mahabadi, J.~Henderson, and S.~Ruder.
\newblock Compacter: Efficient low-rank hypercomplex adapter layers.
\newblock In {\em Proceedings of NeurIPS}, 2021.

\bibitem{mao2022unipelt}
Y.~Mao, L.~Mathias, R.~Hou, A.~Almahairi, H.~Ma, J.~Han, S.~Yih, and M.~Khabsa.
\newblock Unipelt: A unified framework for parameter-efficient language model tuning.
\newblock In {\em Annual Meeting of the Association for Computational Linguistics}, 2022.

\bibitem{pang2022masked}
Y.~Pang, W.~Wang, F.~E. Tay, W.~Liu, Y.~Tian, and L.~Yuan.
\newblock Masked autoencoders for point cloud self-supervised learning.
\newblock In {\em ECCV}, 2022.

\bibitem{pfeiffer-2020-adapterfusion}
J.~Pfeiffer, A.~Kamath, A.~R{\"u}ckl{\'e}, K.~Cho, and I.~Gurevych.
\newblock Adapterfusion: Non-destructive task composition for transfer learning.
\newblock {\em arXiv preprint arXiv:2005.00247}, 2020.

\bibitem{phan2018dgcnn}
A.~V. Phan, M.~Le~Nguyen, Y.~L.~H. Nguyen, and L.~T. Bui.
\newblock Dgcnn: A convolutional neural network over large-scale labeled graphs.
\newblock {\em Neural Networks}, 2018.

\bibitem{Pires2023}
T.~P. Pires, A.~V. Lopes, Y.~Assogba, and H.~Setiawan.
\newblock One wide feedforward is all you need.
\newblock {\em arXiv preprint arXiv:2309.01826}, 2023.

\bibitem{qi2017pointnet}
C.~R. Qi, H.~Su, K.~Mo, and L.~J. Guibas.
\newblock Pointnet: Deep learning on point sets for 3d classification and segmentation.
\newblock In {\em Proceedings of the IEEE conference on computer vision and pattern recognition}, pages 652--660, 2017.

\bibitem{qi2017pointnet++}
C.~R. Qi, L.~Yi, H.~Su, and L.~J. Guibas.
\newblock Pointnet++: Deep hierarchical feature learning on point sets in a metric space.
\newblock In {\em NeurIPS}, 2017.

\bibitem{qi2023recon}
Z.~Qi, R.~Dong, G.~Fan, Z.~Ge, X.~Zhang, K.~Ma, and L.~Yi.
\newblock Contrast with reconstruct: Contrastive 3d representation learning guided by generative pretraining.
\newblock In {\em ICML}, 2023.

\bibitem{qian2022pointnext}
G.~Qian, Y.~Li, H.~Peng, J.~Mai, H.~Hammoud, M.~Elhoseiny, and B.~Ghanem.
\newblock Pointnext: Revisiting pointnet++ with improved training and scaling strategies.
\newblock In {\em NeurIPS}, 2022.

\bibitem{Reid2021}
M.~Reid, E.~Marrese-Taylor, and Y.~Matsuo.
\newblock Subformer: Exploring weight sharing for parameter efficiency in generative transformers.
\newblock {\em arXiv preprint arXiv:2101.00234}, 2021.

\bibitem{renduchintala2023tied}
A.~Renduchintala, T.~Konuk, and O.~Kuchaiev.
\newblock Tied-lora: Enhacing parameter efficiency of lora with weight tying.
\newblock {\em arXiv preprint arXiv:2311.09578}, 2023.

\bibitem{Takase2023}
S.~Takase and S.~Kiyono.
\newblock Lessons on parameter sharing across layers in transformers.
\newblock In N.~Sadat~Moosavi, I.~Gurevych, Y.~Hou, G.~Kim, Y.~J. Kim, T.~Schuster, and A.~Agrawal, editors, {\em Proceedings of The Fourth Workshop on Simple and Efficient Natural Language Processing (SustaiNLP)}, pages 78--90, Toronto, Canada (Hybrid), July 2023. Association for Computational Linguistics.

\bibitem{uy2019revisiting}
M.~A. Uy, Q.-H. Pham, B.-S. Hua, T.~Nguyen, and S.-K. Yeung.
\newblock Revisiting point cloud classification: A new benchmark dataset and classification model on real-world data.
\newblock In {\em Porc. of IEEE Intl. Conf. on Computer Vision}, 2019.

\bibitem{van2008visualizing}
L.~Van~der Maaten and G.~Hinton.
\newblock Visualizing data using t-sne.
\newblock {\em Journal of machine learning research}, 9(11), 2008.

\bibitem{wang2021unsupervised}
H.~Wang, Q.~Liu, X.~Yue, J.~Lasenby, and M.~J. Kusner.
\newblock Unsupervised point cloud pre-training via occlusion completion.
\newblock In {\em ICCV}, 2021.

\bibitem{wang2024prolora}
S.~Wang, B.~Xue, J.~Ye, J.~Jiang, L.~Chen, L.~Kong, and C.~Wu.
\newblock Prolora: Partial rotation empowers more parameter-efficient lora.
\newblock {\em arXiv preprint arXiv:2402.16902}, 2024.

\bibitem{wang2019dynamic}
Y.~Wang, Y.~Sun, Z.~Liu, S.~E. Sarma, M.~M. Bronstein, and J.~M. Solomon.
\newblock Dynamic graph cnn for learning on point clouds.
\newblock {\em ACM Trans. Gr.}, 2019.

\bibitem{wu2024multi}
X.~Wu, S.~Huang, W.~Wang, and F.~Wei.
\newblock Multi-head mixture-of-experts.
\newblock {\em arXiv preprint arXiv:2404.15045}, 2024.

\bibitem{wu20153d}
Z.~Wu, S.~Song, A.~Khosla, F.~Yu, L.~Zhang, X.~Tang, and J.~Xiao.
\newblock 3d shapenets: A deep representation for volumetric shapes.
\newblock In {\em Proc. of IEEE Intl. Conf. on Computer Vision and Pattern Recognition}, 2015.

\bibitem{moec}
Y.~Xie, S.~Huang, T.~Chen, and F.~Wei.
\newblock Moec: Mixture of expert clusters.
\newblock In {\em Proceedings of the AAAI Conference on Artificial Intelligence}, volume~37, pages 13807--13815, 2023.

\bibitem{yu2022point}
X.~Yu, L.~Tang, Y.~Rao, T.~Huang, J.~Zhou, and J.~Lu.
\newblock Point-bert: Pre-training 3d point cloud transformers with masked point modeling.
\newblock In {\em CVPR}, 2022.

\bibitem{bitfit}
E.~B. Zaken, Y.~Goldberg, and S.~Ravfogel.
\newblock Bitfit: Simple parameter-efficient fine-tuning for transformer-based masked language-models.
\newblock In {\em Proceedings of ACL}, 2022.

\bibitem{zha2023instance}
Y.~Zha, J.~Wang, T.~Dai, B.~Chen, Z.~Wang, and S.-T. Xia.
\newblock Instance-aware dynamic prompt tuning for pre-trained point cloud models.
\newblock In {\em Porc. of IEEE Intl. Conf. on Computer Vision}, 2023.

\bibitem{vtab}
X.~Zhai, J.~Puigcerver, A.~Kolesnikov, P.~Ruyssen, C.~Riquelme, M.~Lucic, J.~Djolonga, A.~S. Pinto, M.~Neumann, A.~Dosovitskiy, L.~Beyer, O.~Bachem, M.~Tschannen, M.~Michalski, O.~Bousquet, S.~Gelly, and N.~Houlsby.
\newblock The visual task adaptation benchmark.
\newblock {\em arXiv preprint}, arXiv:1910.04867, 2019.

\bibitem{zhang2022point}
R.~Zhang, Z.~Guo, P.~Gao, R.~Fang, B.~Zhao, D.~Wang, Y.~Qiao, and H.~Li.
\newblock Point-m2ae: multi-scale masked autoencoders for hierarchical point cloud pre-training.
\newblock In {\em NeurIPS}, 2022.

\bibitem{noah}
Y.~Zhang, K.~Zhou, and Z.~Liu.
\newblock Neural prompt search.
\newblock {\em arXiv preprint}, arXiv:2206.04673, 2022.

\bibitem{zhou2024autopeft}
H.~Zhou, X.~Wan, I.~Vuli{\'c}, and A.~Korhonen.
\newblock Autopeft: Automatic configuration search for parameter-efficient fine-tuning.
\newblock {\em Transactions of the Association for Computational Linguistics}, 12:525--542, 2024.

\bibitem{zhou2024dapt}
X.~Zhou, D.~Liang, W.~Xu, X.~Zhu, Y.~Xu, Z.~Zou, and X.~Bai.
\newblock Dynamic adapter meets prompt tuning: Parameter-efficient transfer learning for point cloud analysis.
\newblock {\em arXiv preprint arXiv:2403.01439}, 2024.

\bibitem{zhu2021counter}
Y.~Zhu, J.~Feng, C.~Zhao, M.~Wang, and L.~Li.
\newblock Counter-interference adapter for multilingual machine translation.
\newblock In {\em Proc. of Conf. on Empirical Methods in Natural Language Processing}, 2021.

\end{thebibliography}
}

\newpage
\appendix
\vspace{20pt}

\textbf{{\Large Appendix for \methodshort{}}}

\section{Pytorch-like code of Sharing Multi-head-MoA}
    


\begin{algorithm}[H]
    \caption{PyTorch-like code of Sharing MultiHead-MoA.}
    \label{algo:algo}
    \footnotesize
    \begin{alltt}
    \color{ForestGreen}
\end{alltt}
\begin{alltt}
import torch
import torch.nn as nn
class Sharing_MoA(nn.Module):
    def __init__(self, n_experts, rank, expert, d_model):
        super().__init__()
        self.experts = nn.ModuleList([expert() for _ in range(n_experts)])
        self.switch = nn.Parameter(torch.randn(n_experts, rank))
        self.switch_reduction = nn.Linear(d_model // self.num_heads, 
                                    rank, bias=False)
    def forward(self, x):
        b, l, d = x.shape()
        \color{ForestGreen}# Split input to multi-head \color{Black}
        x = x.view(-1, d_model // self.num_heads).contiguous()
        \color{ForestGreen}# Dimension Reduction \color{Black}
        input = self.switch_reduction(x)
        \color{ForestGreen}# Compute routing probabilities \color{Black}
        route_prob = torch.softmax(self.switch @ input_reduced.T, dim=0)
        \color{ForestGreen}# Route input through experts based on probabilities \color{Black}
        expert_output = torch.stack([expert(x) for expert in self.experts]).T
        \color{ForestGreen}# Same as Switch Transformer \color{Black}
        final_output = torch.bmm(route_prob.unsqueeze(1), expert_output)
                                                                .squeeze(1)
        final_output = final_output.reshape(b, l, d)
        return final_output
class VisionTransformer(nn.Module):
    def __init__(self, num_experts, expert, d_model, depth):
        super().__init__()
        self.adapter = Sharing_MoA(n_experts=num_experts, 
        rank=d_model, 
        expert=expert, 
        d_model=d_model)
        self.layers = nn.ModuleList([Block(adapter=self.adapter, d_model=d_model) 
                                    for _ in range(depth)])
    def forward(self, x):
        for layer in self.layers:
            x = layer(x)
        return x
\end{alltt}
\end{algorithm}














    






\section{Full 2d ablation results on VTAB-1K}
\begin{table*}[h]
\caption{\textbf{Full results on the VTAB-1K benchmark}. }
\vspace{-0.3cm}
\begin{center}
\setlength{\tabcolsep}{0.3pt}
\resizebox{\textwidth}{!}{
\begin{tabular}{p{3cm}<{}p{1.25cm}<{\centering}p{0.75cm}<{\centering}|p{0.75cm}<{\centering}p{0.75cm}<{\centering}p{0.75cm}<{\centering}p{0.75cm}<{\centering}p{0.75cm}<{\centering}p{0.75cm}<{\centering}p{0.75cm}<{\centering}|p{0.75cm}<{\centering}p{0.75cm}<{\centering}p{0.75cm}<{\centering}p{0.75cm}<{\centering}|p{0.75cm}<{\centering}p{0.75cm}<{\centering}p{0.75cm}<{\centering}p{0.75cm}<{\centering}p{0.75cm}<{\centering}p{0.75cm}<{\centering}p{0.75cm}<{\centering}p{0.75cm}<{\centering}}
\toprule
\multicolumn{3}{c|}{}&\multicolumn{7}{c|}{\textbf{Natural}}&\multicolumn{4}{c|}{\textbf{Specialized}}&\multicolumn{8}{c}{\textbf{Structured}}\\
&\multicolumn{1}{c}{\STAB{\rotatebox[origin=c]{90}{Params. (M)}}}
&\multicolumn{1}{c|}{\STAB{\rotatebox[origin=c]{90}{Avg. Acc.}}}
&\multicolumn{1}{c}{\STAB{\rotatebox[origin=c]{90}{Cifar100}}}
&\multicolumn{1}{c}{\STAB{\rotatebox[origin=c]{90}{Caltech101}}}
&\multicolumn{1}{c}{\STAB{\rotatebox[origin=c]{90}{DTD}}}
&\multicolumn{1}{c}{\STAB{\rotatebox[origin=c]{90}{Flower102}}}
&\multicolumn{1}{c}{\STAB{\rotatebox[origin=c]{90}{Pets}}}
&\multicolumn{1}{c}{\STAB{\rotatebox[origin=c]{90}{SVHN}}}
&\multicolumn{1}{c|}{\STAB{\rotatebox[origin=c]{90}{Sun397}}}
&\multicolumn{1}{c}{\STAB{\rotatebox[origin=c]{90}{Camelyon}}}
&\multicolumn{1}{c}{\STAB{\rotatebox[origin=c]{90}{EuroSAT}}}
&\multicolumn{1}{c}{\STAB{\rotatebox[origin=c]{90}{Resisc45}}}
&\multicolumn{1}{c|}{\STAB{\rotatebox[origin=c]{90}{Retinopathy}}}
&\multicolumn{1}{c}{\STAB{\rotatebox[origin=c]{90}{Clevr-Count}}}
&\multicolumn{1}{c}{\STAB{\rotatebox[origin=c]{90}{Clevr-Dist}}}
&\multicolumn{1}{c}{\STAB{\rotatebox[origin=c]{90}{DMLab}}}
&\multicolumn{1}{c}{\STAB{\rotatebox[origin=c]{90}{KITTI-Dist}}}
&\multicolumn{1}{c}{\STAB{\rotatebox[origin=c]{90}{dSpr-Loc}}}
&\multicolumn{1}{c}{\STAB{\rotatebox[origin=c]{90}{dSpr-Ori}}}
&\multicolumn{1}{c}{\STAB{\rotatebox[origin=c]{90}{sNORB-Azim}}}
&\multicolumn{1}{c}{\STAB{\rotatebox[origin=c]{90}{sNORB-Ele}}}\\
\specialrule{0em}{1pt}{1pt}
\hline
\specialrule{0em}{1pt}{1pt}

\specialrule{0em}{1pt}{1pt}

\specialrule{0em}{1pt}{1pt}

\quad\ Both &{0.17}&76.2&73.4&93.3&71.8&99.3&91.4&87.8&56.7&87.5&95.5&84.8&75.8&80.7&62.0&50.5&79.0&85.8&53.7&32.7&41.2\\
\hline
\quad\ Only \\ Sharing-MoA&{1.15}&74.9&69.3&93.5&70.1&99.2&91.3&87.1&55.4&86.9&95.2&84.5&75.2&76.8&61.6&47.1&77.9&82.2&50.1&32.4&38.4\\
\hline
\quad\ Only \\ Block-specific&\bf {1.8}&76.4&73.6&93.1&71.8&99.3&91.0&90.1&56.4&88.2&95.6&86.4&73.7&79.8&62.6&50.2&81.0&84.4&53.5&38.1&38.9\\

\bottomrule
\end{tabular}
}
\end{center}
\label{tab:vtab2}
\vspace{-0.5cm}
\end{table*}



\newpage
\section{Visualization}

\begin{figure*}[h]
\centering
\includegraphics[width=0.65\linewidth]{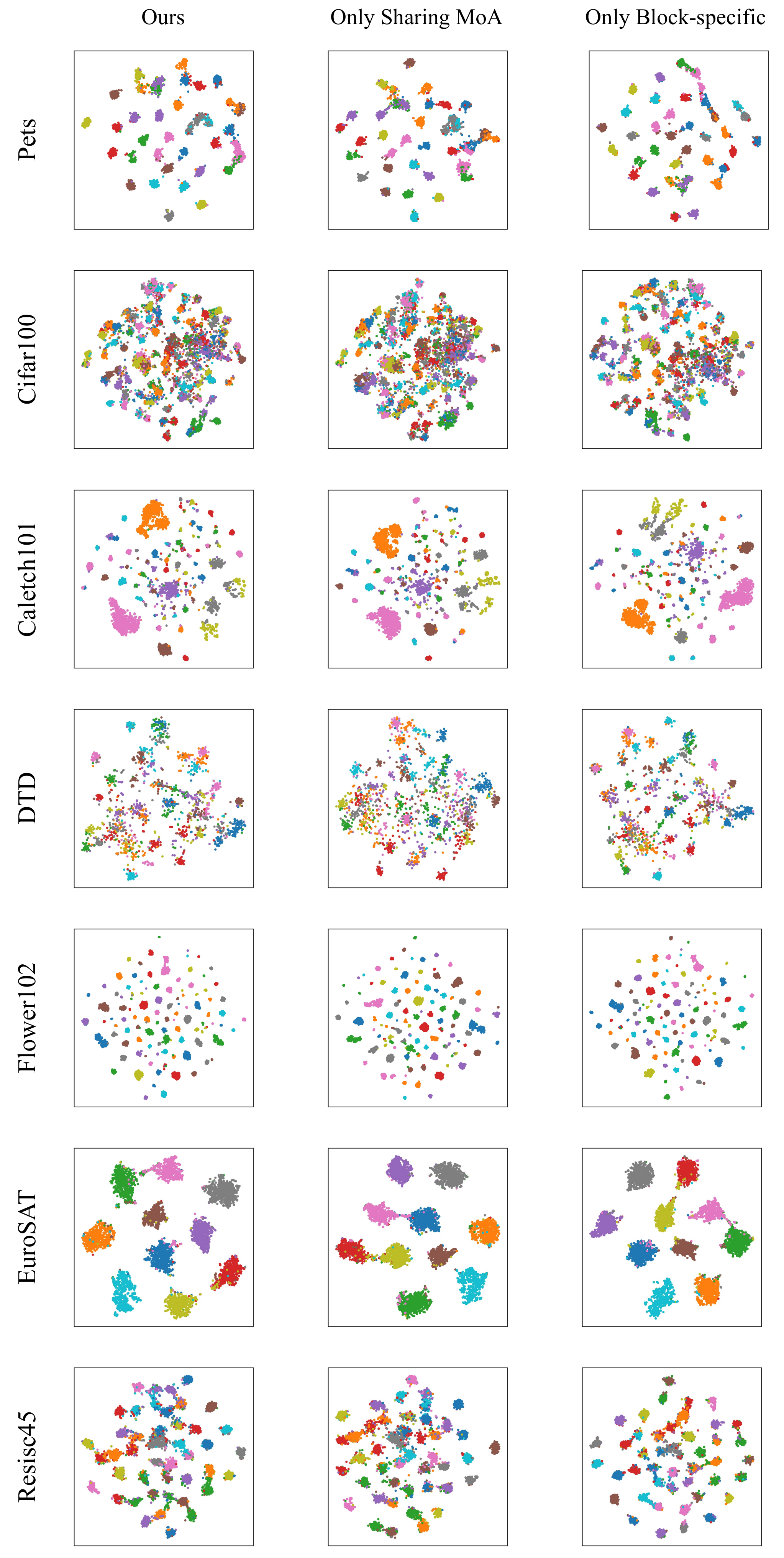}
\caption{More t-SNE visualization results of our ablation study on the 2D VTAB-1K dataset.}
\label{fig:Appendix_visualize_tsne}
\vspace{-0.5cm}
\end{figure*}

\section{Experimental Setting}
\label{sec:appexp_setting}
We follow the implement of the original codebase: NOAH\footnote{\url{https://github.com/ZhangYuanhan-AI/NOAH}}, Bi-adapter\footnote{\url{https://github.com/JieShibo/PETL-ViT/tree/main/binary_adapter}} and DAPT\footnote{\url{https://github.com/LMD0311/DAPT}}. We draw inspiration from Switch Transformer\footnote{\url{https://nn.labml.ai/transformers/switch/index.html}} to implement MoA. The hyperparameters involved in the Sharing MoA module are: num of heads is 3, and num of experts is 4. Reduction dimension keeps the same as the rank of each adapter expert.
\section{Limitations}\label{sec:limitation}
In this work, we only focus on the image and point cloud modalities of visual modalities, because adapters have made more progress in the visual field due to their flexibility. In the future, we will explore how to apply our \methodshort{} method in NLP and even generation tasks.

\end{document}